\documentclass[conference,letterpaper]{IEEEtran}
\IEEEoverridecommandlockouts

\usepackage[T1]{fontenc}
\usepackage[utf8]{inputenc}

\usepackage{newtxtext}
\usepackage{newtxmath}

\usepackage{amsmath,amssymb,amsfonts,mathtools}
\usepackage{bm}

\usepackage{booktabs,multirow}
\usepackage{tabularx}

\usepackage{graphicx}
\graphicspath{{./}{./Figures/}{./figs/}{./Pictures/}}

\usepackage[caption=false,font=footnotesize]{subfig}

\usepackage{enumitem}
\usepackage{xcolor}
\usepackage{siunitx}

\usepackage{placeins}
\usepackage{cite}
\usepackage{stfloats}
\usepackage[hyphens]{url}
\usepackage[hidelinks,breaklinks=true]{hyperref}
\usepackage[nameinlink]{cleveref}

\newcommand{\RPHtext}[1]{\textcolor{red}{\texttt{\detokenize{RESULT_PLACEHOLDER_#1}}}}
\newcommand{\RPH}[1]{\ifmmode \text{\RPHtext{#1}}\else \RPHtext{#1}\fi}

\newif\ifanonymous
\anonymousfalse

\newif\ifshowqual
\showqualtrue

\newcommand{\xr}{x_r}
\newcommand{\xs}{x_s}
\newcommand{\SUT}{F}
\newcommand{\Halg}{H}

\newcommand{\Mmap}{M}
\newcommand{\Smap}{S}

\newcommand{\cvec}{\bm{c}}
\newcommand{\CDFF}{\mathrm{C\text{-}DFF}}

\newcommand{\R}{\mathbb{R}}
\newcommand{\1}{\mathbb{I}}
\newcommand{\eps}{\varepsilon}

\DeclareMathOperator{\Align}{Align}

\newcommand{\Distc}{D_c}

\usepackage[final]{microtype}

\setlist[itemize]{leftmargin=*,noitemsep,topsep=1pt,partopsep=0pt,parsep=0pt}
\setlist[enumerate]{leftmargin=*,noitemsep,topsep=1pt,partopsep=0pt,parsep=0pt}

\AtBeginDocument{%
  \setlength{\abovedisplayskip}{5pt plus 1pt minus 1pt}%
  \setlength{\belowdisplayskip}{5pt plus 1pt minus 1pt}%
  \setlength{\abovedisplayshortskip}{4pt plus 1pt minus 1pt}%
  \setlength{\belowdisplayshortskip}{4pt plus 1pt minus 1pt}%
}
\newcommand{\Npairs}{\num{2126}}

\newcommand{\MaybeIncludeGraphics}[2][]{%
  \IfFileExists{#2}{%
    \includegraphics[#1]{#2}%
  }{%
    \fbox{\parbox{0.95\linewidth}{\centering
      \textbf{[Figure Placeholder]}\\
      \texttt{#2}\\[3pt]
      Replace with the final rendered figure file.}}%
  }%
}

\makeatletter
\let\oldthebibliography\thebibliography
\let\endoldthebibliography\endthebibliography
\renewenvironment{thebibliography}[1]{%
  \oldthebibliography{#1}%
  \setlength{\itemsep}{0pt}%
  \setlength{\parskip}{0pt}%
  \setlength{\parsep}{0pt}%
}{%
  \endoldthebibliography%
}
\makeatother

\title{C-DFF: Concept-Level Diagnostics for Mechanism Parity in Simulation-Based Validation of Autonomous Driving Perception}

\IEEEoverridecommandlockouts

\author{%
\IEEEauthorblockN{%
Danial Safaei\IEEEauthorrefmark{1},
Siddartha Khastgir\IEEEauthorrefmark{1},
Matthew Higgins\IEEEauthorrefmark{1},
Jeroen Ploeg\IEEEauthorrefmark{3},
Mohsen Alirezaei\IEEEauthorrefmark{2},\\
Son Tong\IEEEauthorrefmark{4} and
Ludo Gielen\IEEEauthorrefmark{4}%
}

\IEEEauthorblockA{%
\IEEEauthorrefmark{1}WMG, University of Warwick, Coventry, U.K.\\
\texttt{\{danial.safaei, s.khastgir.1, m.higgins\}@warwick.ac.uk}%
}

\IEEEauthorblockA{%
\IEEEauthorrefmark{2}Eindhoven University of Technology (TU/e), Eindhoven, The Netherlands\\
\texttt{m.alirezaei@tue.nl}%
}

\IEEEauthorblockA{%
\IEEEauthorrefmark{3}2getthere, Utrecht, The Netherlands\\
\texttt{jeroen@2getthere.eu}%
}

\IEEEauthorblockA{%
\IEEEauthorrefmark{4}Siemens Digital Industries Software, Leuven, Belgium\\
\texttt{\{son.tong, ludo.gielen\}@siemens.com}%
}

\thanks{Corresponding author: Danial Safaei.}%
}

\newif\ifarxivversion
\arxivversiontrue
\ifarxivversion
\usepackage{eso-pic}
\AddToShipoutPictureBG*{%
  \AtPageUpperLeft{\raisebox{-30pt}{\makebox[\paperwidth][c]{%
    \colorbox{yellow!25}{\parbox[t]{\textwidth}{\centering\footnotesize
      \textbf{Accepted and presented at the 2026 IEEE 29th International Conference on Intelligent Transportation Systems (ITSC),}\\
      \textbf{Naples, Italy, September 15--18, 2026.} This is the accepted version of the paper.}}}}}%
  \AtPageLowerLeft{\raisebox{24pt}{\makebox[\paperwidth][c]{%
    \parbox[b]{\textwidth}{\centering\scriptsize
      \copyright\,2026 IEEE. Personal use of this material is permitted. Permission from IEEE must be obtained for all other uses, in any current or future media, including reprinting/republishing this material for advertising or promotional purposes, creating new collective works, for resale or redistribution to servers or lists, or reuse of any copyrighted component of this work in other works.}}}}%
}
\fi

\begin{document}
\maketitle
\thispagestyle{empty}
\pagestyle{empty}

\begin{abstract}
Simulation is widely used in autonomous-driving validation, but matching task outputs between real and synthetic images does not guarantee that a system relies on the same visual evidence.
This creates a validation gap for transfer, especially in safety-critical perception functions.

We present \emph{Concept-Decisive Feature Fidelity} (C-DFF), a concept-level mechanism-parity diagnostic for paired real--synthetic evaluation.
Starting from decisive evidence extracted from a frozen system-under-test (SUT), C-DFF projects that evidence into a shared semantic concept space using off-the-shelf segmentation and compares real and synthetic concept-importance vectors.
The result is not only a scalar mechanism-parity score, but also a concept-shift diagnostic that identifies which scene concepts become more or less decisive in simulation.

We evaluate C-DFF on KITTI $\leftrightarrow$ Virtual~KITTI~2 with \Npairs{} matched pairs across three frozen SUTs (steering regression, YOLOP drivable-area, YOLOP lane-line).
Across these tasks, C-DFF is SUT-discriminative and yields interpretable concept-level summaries that support validation triage and simulator debugging.
When reported alongside task-specific output-value context metrics, C-DFF shows that output agreement does not by itself establish mechanism parity.
These results position C-DFF as a practical semantic diagnostic layer for validation and verification workflows in simulation-based AV testing.
\end{abstract}


\section{Introduction}
\label{sec:intro}

Simulation is now a core component of autonomous-driving development and safety evaluation because it enables scalable testing across scenarios that are rare, risky, or expensive to reproduce on-road \cite{Riedmaier2020Survey,Zhong2021ScenarioSurvey,ADWGSTV2Whitepaper}.
However, simulation-based validation still faces a key question: when a real scene and its synthetic counterpart produce similar model outputs, did the system rely on the same evidence to make that decision?
This question matters because output agreement alone can hide ``right output for the wrong reason'' behavior.
A perception or driving model may preserve the final prediction while shifting its reliance toward spurious or non-transferable cues, which weakens the validity of simulation-based conclusions \cite{Lapuschkin2019CleverHans,Geirhos2020Shortcut}.
For validation and verification workflows, practitioners therefore need metrics that test not only \emph{output similarity}, but also \emph{mechanism parity}.
Prior work introduced Decisive Feature Fidelity (DFF) as a way to compare decisive evidence maps extracted from a frozen system-under-test (SUT) \cite{Safaei2025DFF}.
DFF captures mechanism-level differences that output metrics can miss, but raw decisive-map differences are still difficult to translate into engineering action.
A heatmap mismatch indicates that the mechanism changed, but it does not directly reveal which semantic parts of the scene became more or less decisive.
This paper addresses that interpretability and actionability gap.
We introduce \emph{Concept-Decisive Feature Fidelity} (C-DFF), a concept-level mechanism-parity diagnostic for paired real--synthetic evaluation.
C-DFF combines decisive evidence extraction on a frozen SUT with off-the-shelf semantic segmentation to project decisive evidence into a shared concept vocabulary (e.g., road, sidewalk, building, sky).
The output is a scalar mechanism-parity score together with concept-shift summaries that are directly usable for simulator, renderer, and synthetic-data debugging.
We evaluate C-DFF on KITTI $\leftrightarrow$ Virtual~KITTI~2 using \Npairs{} matched real--synthetic pairs and three frozen SUTs (steering regression, YOLOP drivable-area, YOLOP lane-line).
Across these tasks, C-DFF is SUT-discriminative.
Reported alongside task-specific output-value context metrics, it shows that output agreement does not by itself establish mechanism parity and therefore captures a different aspect of real--synthetic transfer.
The concept-level diagnostics further identify which scene concepts are responsible for the mismatch, which makes the metric operational for simulation-based validation and verification.

\begin{figure*}[t]
\centering
\MaybeIncludeGraphics[width=0.94\textwidth]{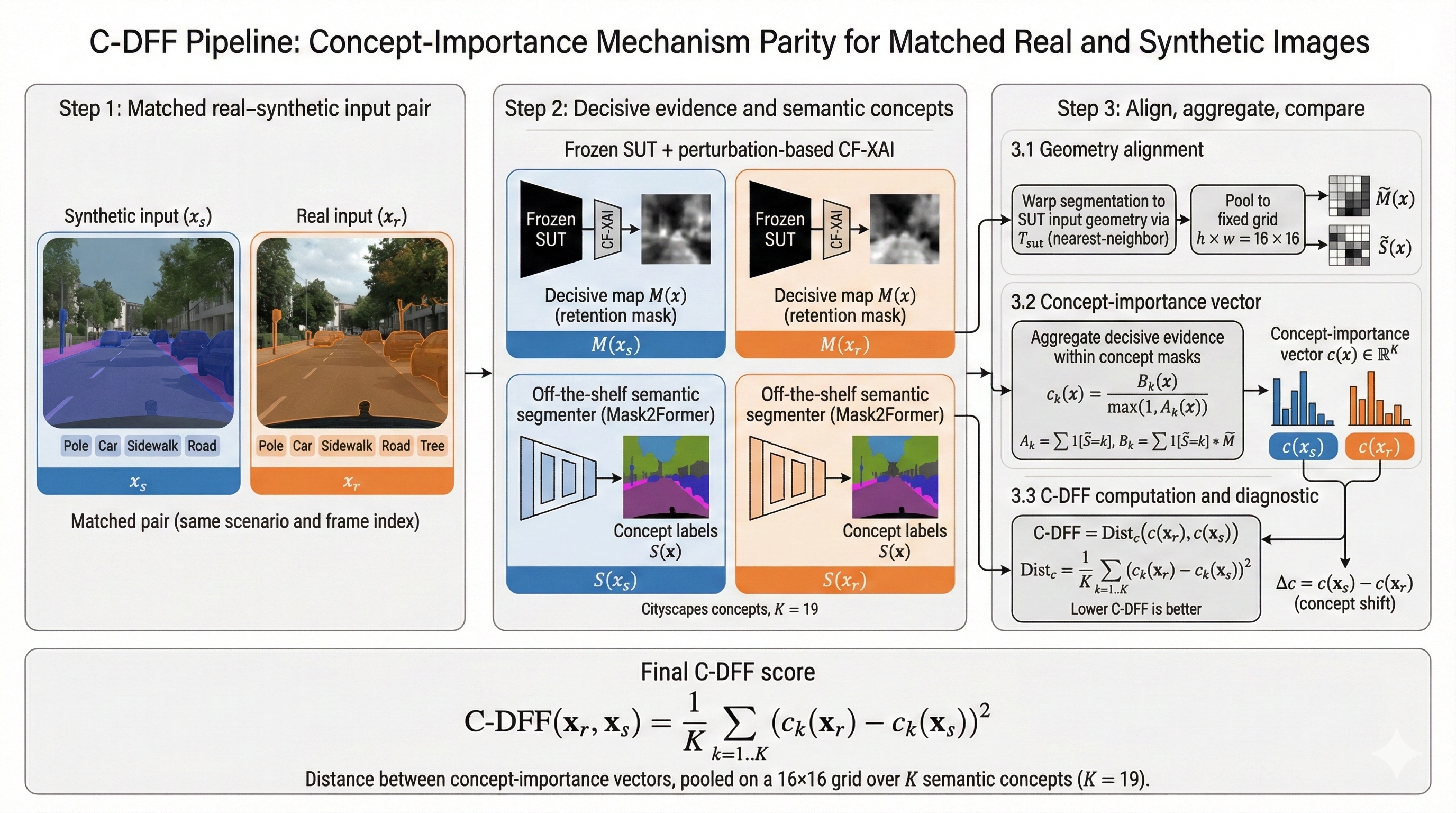}
\caption{C-DFF pipeline. For an image $x$, compute (1) a decisive map $\Mmap(x)=\Halg(\SUT,x)$ on the frozen SUT, (2) semantic segmentation $\Smap(x)$ via an off-the-shelf segmenter, (3) aligned maps $\tilde{\Mmap}(x),\tilde{\Smap}(x)$ by applying the SUT geometry and then pooling to a compact shared grid, using average pooling for decisive maps and nearest-neighbor resizing for labels, (4) a concept-importance vector $\cvec(x)$ by aggregating decisive evidence within semantic concept masks, and (5) C-DFF as a distance between $\cvec(\xr)$ and $\cvec(\xs)$.}
\label{fig:pipeline}
\end{figure*}

\paragraph*{Position relative to DFF}
This paper is deliberately scoped differently from DFF.
DFF asks whether decisive evidence maps match and uses that signal as part of a calibration loop for synthetic-data generation \cite{Safaei2025DFF}.
C-DFF addresses a different problem: once a mechanism mismatch exists, how can it be expressed in a semantic form that is interpretable enough for validation triage and debugging?
Accordingly, this paper does \emph{not} introduce a generator-calibration objective.
Its contribution is a concept-space diagnostic layer that converts decisive evidence into concept-level mismatch summaries.

Current V\&V practice often relies on output-value agreement between real and synthetic runs, but such agreement can still hide ``right output for the wrong reason'' failures.
Recent work therefore treats fidelity as SUT-dependent, including instance-level views of safety-aware synthetic fidelity \cite{Zhang2018LPIPS,Cheng2024SAFidelity,ADWGSTV2Whitepaper}.
C-DFF preserves this mechanism-parity objective while expressing decisive evidence in a shared concept vocabulary.
We extract decisive evidence with perturbation-based explanations and aggregate it over segmenter-defined scene concepts, giving an interpretable semantic scaffold without retraining the SUT \cite{Adebayo2018Sanity,FongVedaldi2017MeaningfulPerturbation,Kim2018TCAV,Koh2020CBM,Cordts2016Cityscapes,Cheng2022Mask2Former}.

\subsection{Contributions}
\label{sec:contrib}

\begin{itemize}
\item \textbf{Concept-space mechanism-parity formulation.}
We define C-DFF, which compares matched real--synthetic inputs through distances between concept-importance vectors obtained by aggregating decisive evidence over semantic scene concepts.
\item \textbf{Actionable concept-shift diagnostics.}
Beyond a scalar score, we define per-pair and dataset-level concept-shift summaries that identify which concepts become more or less decisive in synthetic imagery.
\item \textbf{Validation-oriented operationalization.}
We show how C-DFF can be used as a scalar gate plus a ranked semantic diagnostic for screening, triage, and requirement-style monitoring in simulation-based validation workflows.
\item \textbf{Multi-SUT empirical study.}
We report C-DFF on \Npairs{} matched KITTI $\leftrightarrow$ Virtual~KITTI~2 pairs across steering, drivable-area, and lane-line SUTs, and report task-specific output-value context metrics to show that output agreement need not imply mechanism parity.
\end{itemize}

The remainder of the paper is organized as follows.
Section~\ref{sec:setup_problem} defines the paired real--synthetic setting and frozen SUT regime.
Section~\ref{sec:method} specifies C-DFF and its diagnostics.
Sections~\ref{sec:exp_setup} and \ref{sec:results} describe the evaluation setup and report results.
Section~\ref{sec:discussion} discusses operational use and threats to validity, and Section~\ref{sec:conclusion} concludes.

\section{Problem Formulation}
\label{sec:setup_problem}

\paragraph*{Paired real--synthetic dataset}
We assume a paired dataset
\[
\mathcal{D}=\{(\xr^{(i)},\xs^{(i)})\}_{i=1}^{N},
\]
where $\xr^{(i)}\in\R^{H\times W\times 3}$ is a real image and $\xs^{(i)}$ is a synthetic image intended to represent the same underlying scenario as $\xr^{(i)}$.
For cloned data such as KITTI $\leftrightarrow$ Virtual~KITTI~2, this means matching viewpoint, sequence identity, and frame index as closely as the benchmark permits.
Pairing quality is a first-order requirement because misalignment inflates mechanism-parity distances.

\paragraph*{Frozen systems-under-test}
Let $\SUT_j:\R^{H\times W\times 3}\rightarrow\R^{d_j}$ be the $j$-th SUT, for $j=1,\dots,J$.
We evaluate each SUT in a strictly frozen regime: weights and preprocessing are fixed.
C-DFF is used for assessment and diagnosis, not for training or adapting the SUT.

\paragraph*{Goal: concept-level mechanism parity}
For each frozen SUT and each matched real--synthetic pair $(\xr,\xs)$, we measure whether the SUT relies on similar \emph{semantic} evidence across domains.
The output is a scalar mechanism-parity score and a concept-shift diagnostic that explains which scene concepts changed in importance.

\section{Method: Concept-Decisive Feature Fidelity (C-DFF)}
\label{sec:method}

Figure~\ref{fig:pipeline} summarizes the C-DFF workflow.
Starting from a frozen SUT and a paired image, we first estimate decisive evidence, then project that evidence into a semantic concept space using off-the-shelf segmentation, and finally compare real and synthetic concept-importance vectors.
We now define each step formally.

\subsection{Decisive evidence maps}
\label{sec:method_decisive}

Let $T_{\mathrm{sut}}$ denote the fixed deterministic SUT preprocessing transform (e.g., resize and crop to the SUT input resolution).
We define the decisive map on the SUT input geometry:
\[
x_{\mathrm{sut}} \coloneqq T_{\mathrm{sut}}(x),
\qquad
\Mmap(x)\coloneqq \Halg(\SUT,x_{\mathrm{sut}})\in\R_{\ge 0}^{H_{\mathrm{sut}}\times W_{\mathrm{sut}}}.
\]

\paragraph*{Perturbation-based CF-XAI instantiation}
In this paper, we instantiate $\Halg$ with a direct image-space perturbation objective.
Starting from $x_{\mathrm{cf}}^{(0)}=x_{\mathrm{sut}}$, we optimize a counterfactual image so that a task-dependent scalar summary $s(\cdot)$ is driven toward a target-side value $s_{\mathrm{tgt}}$, while keeping the perturbation sparse and smooth:
\begin{equation}
\mathcal{J}_t(x_{\mathrm{cf}})\coloneqq
\bigl(s(x_{\mathrm{cf}})-s_{\mathrm{tgt}}\bigr)^2
+ w_t \,\mathrm{mean}\!\left(\left|x_{\mathrm{cf}}-x_{\mathrm{sut}}\right|\right)
+ \lambda_{\mathrm{TV}}\,\mathrm{TV}(x_{\mathrm{cf}}).
\label{eq:cf_obj}
\end{equation}
We then define $\Mmap(x)$ as the channel-mean absolute perturbation after optimization.
Here $w_t$ controls the perturbation-size penalty and $\lambda_{\mathrm{TV}}$ controls spatial smoothness.
The optimization is used only to extract a diagnostic map for the fixed SUT; it does not update SUT weights.
We use Adam with clamping to $[0,1]$ \cite{KingmaBa2015Adam}, and the TV regularizer follows \cite{Rudin1992TV}.

\paragraph*{Scope note}
The decisive-map extractor is treated here as a modular component of the C-DFF pipeline.
Unlike the DFF paper, which instantiated decisive evidence via a mask-and-infill counterfactual estimator with pooled decisive maps \cite{Safaei2025DFF}, the present paper uses a direct perturbation-based objective because the scientific focus here is \emph{semantic diagnosis}, not generator calibration or estimator comparison.

\paragraph*{Task-specific summaries}
For Steering, $s(x)$ is the scalar steering output.
For YOLOP-DA and YOLOP-LL, $s(x)$ is the mean probability mask of the selected head (softmax for 2-channel outputs, sigmoid for 1-channel outputs).
For each image and SUT, we run one CF optimization and use the resulting decisive map for concept aggregation.

\subsection{Semantic concept segmentation}
\label{sec:method_seg}

We use an off-the-shelf semantic segmenter to obtain a per-pixel concept label map
\[
\Smap(x)\in\{1,\dots,K\}^{H\times W},
\]
where label $k$ corresponds to a semantic concept.
A concrete instantiation uses Mask2Former \cite{Cheng2022Mask2Former} with a Cityscapes-style taxonomy \cite{Cordts2016Cityscapes}.
In this work, we use the Cityscapes-19 taxonomy ($K=19$).
Segmentation is used only as an interpretability scaffold and is not trained or tuned in this work.

\subsection{Deterministic alignment and pooling}
\label{sec:method_align}

A key requirement is that concept masks and decisive evidence refer to the same geometry that the SUT consumed.
Let $T_{\mathrm{sut}}$ be the deterministic SUT preprocessing transform.
We warp the concept labels through the same transform using nearest-neighbor interpolation:
\[
\Smap_{\mathrm{sut}}(x)\coloneqq T_{\mathrm{sut}}\bigl(\Smap(x)\bigr)\in\{1,\dots,K\}^{H_{\mathrm{sut}}\times W_{\mathrm{sut}}}.
\]
We then align $\Mmap(x)$ and $\Smap_{\mathrm{sut}}(x)$ onto a fixed pooling grid of size $h\times w$:
\[
(\tilde{\Mmap}(x),\tilde{\Smap}(x)) = \Align\bigl(\Mmap(x),\Smap_{\mathrm{sut}}(x)\bigr).
\]

\textbf{Primary choice.}
We use $h=16$ and $w=16$ as a compact common support across the different SUT input resolutions and output geometries.
We average-pool $\Mmap(x)$ to $h\times w$ and resize $\Smap_{\mathrm{sut}}(x)$ to $h\times w$ using nearest-neighbor interpolation to preserve discrete labels.
This yields
\[
\tilde{\Mmap}(x)\in\R_{\ge 0}^{h\times w},
\qquad
\tilde{\Smap}(x)\in\{1,\dots,K\}^{h\times w}.
\]

\subsection{Concept-importance vector}
\label{sec:method_conceptvec}

Define the index set of pooled pixels as $\Omega=\{1,\dots,h\}\times\{1,\dots,w\}$.
For each concept $k\in\{1,\dots,K\}$, define its pooled decisive mass and pooled area:
\[
\begin{aligned}
A_k(x)&\coloneqq \sum_{p\in\Omega}\1[\tilde{\Smap}(x)_p=k],\\
B_k(x)&\coloneqq \sum_{p\in\Omega}\1[\tilde{\Smap}(x)_p=k]\cdot\tilde{\Mmap}(x)_p.
\end{aligned}
\]
The area-normalized concept importance is
\begin{equation}
c_k(x)\coloneqq \frac{B_k(x)}{\max(1, A_k(x))}.
\label{eq:ck}
\end{equation}
Stacking all concepts yields the concept-importance vector
\[
\cvec(x)\coloneqq [c_1(x),\dots,c_K(x)]^\top\in\R_{\ge 0}^{K}.
\]
Unless stated otherwise, we use Eq.~\eqref{eq:ck} as defined above and do \emph{not} apply any further across-concept normalization.
Thus, $c_k(x)$ is the average decisive mass assigned to concept $k$ on the pooled grid.
A larger $c_k(x)$ means that pixels belonging to concept $k$ carry more of the perturbation evidence needed to change the fixed SUT output for that image.

\subsection{C-DFF distance between real and synthetic}
\label{sec:method_cdff}

Given a matched pair $(\xr,\xs)$ and a frozen SUT $\SUT$, define
\begin{equation}
\CDFF(\xr,\xs;\SUT)\coloneqq \Distc\bigl(\cvec(\xr),\cvec(\xs)\bigr),
\label{eq:cdff_def}
\end{equation}
where $\Distc$ is a distance on $\R^K$.
Our primary distance is mean-squared error:
\begin{equation}
\Distc^{\mathrm{MSE}}(\cvec,\cvec') \coloneqq \frac{1}{K}\sum_{k=1}^K (c_k-c_k')^2.
\label{eq:dist_mse}
\end{equation}
We use MSE because both inputs are fixed-length, nonnegative vectors on the same semantic concept basis.
This choice is symmetric, preserves magnitude information, and penalizes larger concept-level changes more strongly.

\subsection{Thresholding and pass-rate}
\label{sec:method_thresholds}

For a given frozen SUT, we calibrate descriptive thresholds $\eps_{90},\eps_{95}$ as the 90th and 95th percentiles of C-DFF on a calibration split, then report the held-out pass rate as the fraction of test pairs with $\CDFF(\xr,\xs)\le \eps$.
These thresholds are operational summaries for a fixed split, not safety guarantees.

\subsection{Diagnostics: concept shift vectors}
\label{sec:method_diag}

A scalar C-DFF indicates mismatch magnitude, but not its cause.
We therefore define a per-pair concept-shift vector:
\begin{equation}
\Delta\cvec(\xr,\xs)\coloneqq \cvec(\xs)-\cvec(\xr)\in\R^{K}.
\label{eq:delta_c}
\end{equation}
Interpretation:
$\Delta c_k>0$ indicates concept $k$ is more decisive in synthetic than in real, whereas $\Delta c_k<0$ indicates it is less decisive.

We report the following diagnostic summaries:
\begin{itemize}
\item \textbf{Top-$k$ increases and decreases:} the largest positive and negative components of $\Delta\cvec$ per SUT.
\item \textbf{Dataset-level mean shift:} $\bar{\Delta\cvec}=\frac{1}{N}\sum_{i=1}^N \Delta\cvec(\xr^{(i)},\xs^{(i)})$.
\item \textbf{Concept-level envelopes:} per-concept quantiles of $\Delta c_k$ to support requirement-style checks.
\end{itemize}

\section{Experimental Setup}
\label{sec:exp_setup}

\subsection{Data and frozen SUTs}
\label{sec:data_suts}

We evaluate C-DFF on KITTI $\leftrightarrow$ Virtual~KITTI~2 \cite{Geiger2013KITTI,Cabon2020VKITTI2}.
Pairs are constructed at frame level by shared sequence ID and frame index (e.g., KITTI tracking sequence 0001$\to$V-KITTI Scene01).
From the matched corpus, we report results on \Npairs{} paired frames.
The evaluated set covers five paired sequences (Scene01/02/06 for calibration, Scene18/20 for held-out evaluation) with a front-facing monocular camera at $1248\times376$.
Pairing noise and its effect on C-DFF are treated explicitly as a validity threat.

We evaluate exactly the following frozen SUTs:
\begin{enumerate}
\item \textbf{Steering regression} (PilotNet-style head) \cite{Bojarski2016E2E}:
CNN regression model (3-conv + 2-FC layers) predicting a scalar steering angle from $200\times66$ RGB input, trained on the Udacity steering-angle dataset (Challenge 2) \cite{Udacity2016Challenge2,Chernikova2019Udacity}.
\item \textbf{YOLOP drivable-area head (DA)}:
YOLOP multi-task network with shared encoder and task head for drivable-area segmentation, operating on $640\times384$ RGB input \cite{Wu2022YOLOP}.
\item \textbf{YOLOP lane-line head (LL)}:
YOLOP task head for lane-line detection, same backbone family \cite{Wu2022YOLOP}.
\end{enumerate}
All SUT weights are frozen.
Input preprocessing (the $T_{\mathrm{sut}}$ used in Section~\ref{sec:method_decisive}) is as follows:
Steering: resize to $200\times66$, normalize to $[0,1]$.
YOLOP: resize to $640\times384$, normalize to $[0,1]$.
No augmentation at inference.

\subsection{Implementation details}
\label{sec:cf_config}

\paragraph*{CF-XAI configuration}
We instantiate $\Halg$ with direct image-space perturbation optimization (Eq.~\ref{eq:cf_obj}) in the spirit of meaningful perturbations \cite{FongVedaldi2017MeaningfulPerturbation}.
For each image and SUT, we run one CF optimization and convert the optimized perturbation into a decisive map by channel-mean absolute difference.
We use fixed per-SUT optimization settings selected on a tuning subset disjoint from the held-out evaluation split (Optuna, 20 trials per SUT) \cite{Akiba2019Optuna}.
The primary settings are:
80 steps for Steering and 60 steps for YOLOP heads; Adam optimization with clamping to $[0,1]$; target-side margin $\tau=0.1$; total-variation weight $\lambda_{\mathrm{TV}}=\num{1e-3}$; and pooled decisive maps at $16\times16$ for concept aggregation.
The perturbation-size weight is scheduled as $w_t=0.1$ for the first half of the optimization and then increased to $w_t=0.5$ for Steering and $w_t=0.3$ for the YOLOP heads.
For implementation diagnostics, residual convergence is treated relative to the target-side margin using $0.8\tau=0.08$.
These settings are fixed before held-out evaluation.

\paragraph*{Semantic segmenter configuration}
We use an off-the-shelf segmenter, with a concrete default being Mask2Former \cite{Cheng2022Mask2Former}.
The configuration is:
(1) model checkpoint trained on Cityscapes \cite{HFMask2FormerCityscapes},
(2) Cityscapes-19 taxonomy ($K=19$) \cite{Cordts2016Cityscapes},
(3) geometry consistency by warping $\Smap(x)$ through the exact SUT transform $T_{\mathrm{sut}}$ via nearest-neighbor interpolation before pooling (Section~\ref{sec:method_align}).
The segmenter is not trained or adapted here.

\subsection{Metrics reported}
\label{sec:metrics_reported}

\paragraph*{Primary metric}
C-DFF as defined in Eq.~\eqref{eq:cdff_def}, lower is better.

\paragraph*{Context metrics}
We report task-specific output-value similarity metrics for interpretive context:
\begin{itemize}
\item steering: $\exp(-|\theta_r-\theta_s|/0.2)$, where $\theta$ is the predicted steering angle,
\item YOLOP-DA: IoU between binarized drivable-area masks,
\item YOLOP-LL: $0.4\cdot\text{Chamfer}_{\text{score}} + 0.6\cdot F_1^{\text{tol}=3\text{px}}$ on binarized lane-line masks \cite{Borgefors1988Chamfer}.
\end{itemize}
These context metrics are not mechanism-parity metrics.
Because they are defined differently across tasks, their absolute values are not intended for cross-SUT ranking.
They are reported only as within-SUT context to show that output agreement and mechanism parity can diverge.

\section{Results}
\label{sec:results}

\subsection{Main quantitative result}
\label{sec:results_main}

Table~\ref{tab:main_results} reports the main quantitative result.
C-DFF separates the three frozen SUTs on the held-out split.
Because the output-value (OV) context metric is defined differently for each task, its absolute values are not intended for cross-SUT ranking.
Instead, OV is reported as within-SUT context, showing that output agreement does not by itself establish mechanism parity.
Table~\ref{tab:cf_valid} reports implementation-level CF-XAI diagnostics for a held-out synthetic-side subset.

\begin{table}[t]
\centering
\footnotesize
\caption{CF-XAI valid-run diagnostics on the synthetic-side held-out cache subset (50 runs per SUT). ``Valid'' means the optimization completed without numerical failure and produced a finite decisive map. The residual is reported only as a secondary convergence diagnostic; this subset-level diagnostic does not replace the full held-out C-DFF results.}
\label{tab:cf_valid}
\setlength{\tabcolsep}{6pt}
\renewcommand{\arraystretch}{1.02}
\begin{tabular}{@{}lcc@{}}
\toprule
\textbf{SUT} & \textbf{Valid-run rate} & \textbf{Median }$\lvert s(x_{\mathrm{cf}}^\star)-s_{\mathrm{tgt}}\rvert$ \\
\midrule
Steering  & 100.0\% & \num{4.04e-1} \\
YOLOP--DA & 100.0\% & \num{1.45e-3} \\
YOLOP--LL & 100.0\% & \num{8.83e-3} \\
\bottomrule
\end{tabular}
\end{table}

\begin{table}[t]
\centering
\footnotesize
\caption{Main C-DFF results on the held-out evaluation split. Lower C-DFF is better. Thresholds $\eps_{90},\eps_{95}$ are calibrated on the calibration split as percentiles of C-DFF. Pass rates are computed on the held-out split. OV is a task-specific output-value context metric, not a mechanism-parity metric, and is not intended for cross-SUT ranking.}
\label{tab:main_results}
\setlength{\tabcolsep}{2.8pt}
\renewcommand{\arraystretch}{1.12}
\begin{tabular}{@{}lcccccc@{}}
\toprule
\textbf{SUT} &
\textbf{Mean} &
\textbf{Med.} &
\textbf{Std.} &
\textbf{P@$\eps_{90}$} &
\textbf{P@$\eps_{95}$} &
\textbf{OV (ctx.)} \\
\midrule
Steering  & \num{3.12e-4} & \num{2.94e-4} & \num{1.71e-4} & 77.1\% & 92.8\% & \num{0.3742} \\
YOLOP--DA & \num{1.25e-4} & \num{1.00e-4} & \num{1.03e-4} & 97.0\% & 100.0\% & \num{0.4518} \\
YOLOP--LL & \num{3.90e-5} & \num{3.20e-5} & \num{3.30e-5} & 66.9\% & 74.7\% & \num{0.0730} \\
\bottomrule
\end{tabular}
\end{table}

\subsection{Diagnostics and additional analyses}
\label{sec:results_diag}

\paragraph*{Bounding segmenter-induced shifts using VKITTI2 ground truth}
Virtual~KITTI~2 provides per-pixel semantic ground truth.
To bound segmentation-induced concept shifts, we recompute $\cvec(\xs)$ on the synthetic side using Mask2Former predictions and VKITTI2 ground-truth labels while keeping $\Mmap(\xs)$ fixed.
Because VKITTI2 labels are not Cityscapes-native, we map VKITTI classes to nearest Cityscapes concepts and ignore unmapped labels before aggregation.
This limited one-sided diagnostic uses synthetic-side labels only, covers the held-out cache subset, and depends on an approximate VKITTI2-to-Cityscapes label mapping.

\paragraph*{Thresholds and distributional view}

\begin{table}[t]
\centering
\scriptsize
\caption{Synthetic-side label-source diagnostic and calibrated C-DFF thresholds. Label-source C-DFF compares Mask2Former predictions against VKITTI2 ground-truth labels on the held-out cache subset. Thresholds are calibrated on the calibration split.}
\label{tab:seg_thresholds}
\setlength{\tabcolsep}{3.2pt}
\renewcommand{\arraystretch}{1.00}
\begin{tabular}{@{}lcccc@{}}
\toprule
\textbf{SUT} &
\textbf{Pred labels} &
\textbf{GT labels} &
$\boldsymbol{\eps_{90}}$ &
$\boldsymbol{\eps_{95}}$ \\
\midrule
Steering  & \num{2.87e-4} & \num{2.41e-4} & \num{4.06e-4} & \num{5.48e-4} \\
YOLOP--DA & \num{7.11e-5} & \num{6.05e-5} & \num{3.99e-4} & \num{5.23e-4} \\
YOLOP--LL & \num{2.28e-5} & \num{2.52e-5} & \num{4.40e-5} & \num{5.10e-5} \\
\bottomrule
\end{tabular}
\end{table}

Table~\ref{tab:seg_thresholds} reports the synthetic-side label-source diagnostic and the calibrated thresholds.
For transparency, we also report held-out percentiles $q_{90}^{\mathrm{test}}$ and $q_{95}^{\mathrm{test}}$ per SUT:
$q_{90}^{\mathrm{test}}=(\text{Steering: }\num{4.36e-4},\ \text{YOLOP--DA: }\num{1.67e-4},\ \text{YOLOP--LL: }\num{4.53e-5})$,
$q_{95}^{\mathrm{test}}=(\text{Steering: }\num{7.17e-4},\ \text{YOLOP--DA: }\num{2.81e-4},\ \text{YOLOP--LL: }\num{5.18e-5})$.

\begin{figure}[t]
\footnotesize
\centering
\MaybeIncludeGraphics[width=0.86\columnwidth]{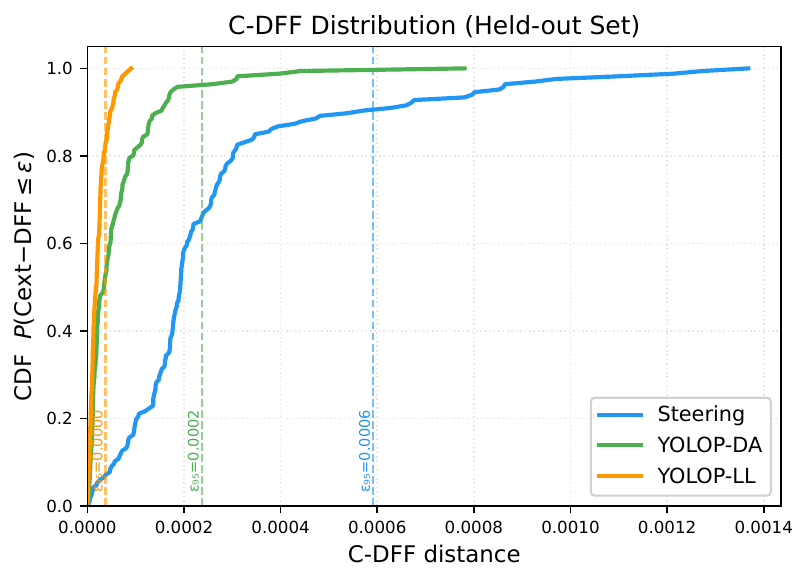}
\caption{C-DFF distributions by SUT on the held-out split. Empirical CDFs $P(\CDFF\le \eps)$ are shown for each SUT; dashed vertical lines indicate calibration-split $\eps_{95}$ values from Table~\ref{tab:seg_thresholds}, not universal safety bounds.}
\label{fig:cdf}
\end{figure}

\begin{figure*}[!b]
\centering
\MaybeIncludeGraphics[
  width=0.94\textwidth,
  keepaspectratio
]{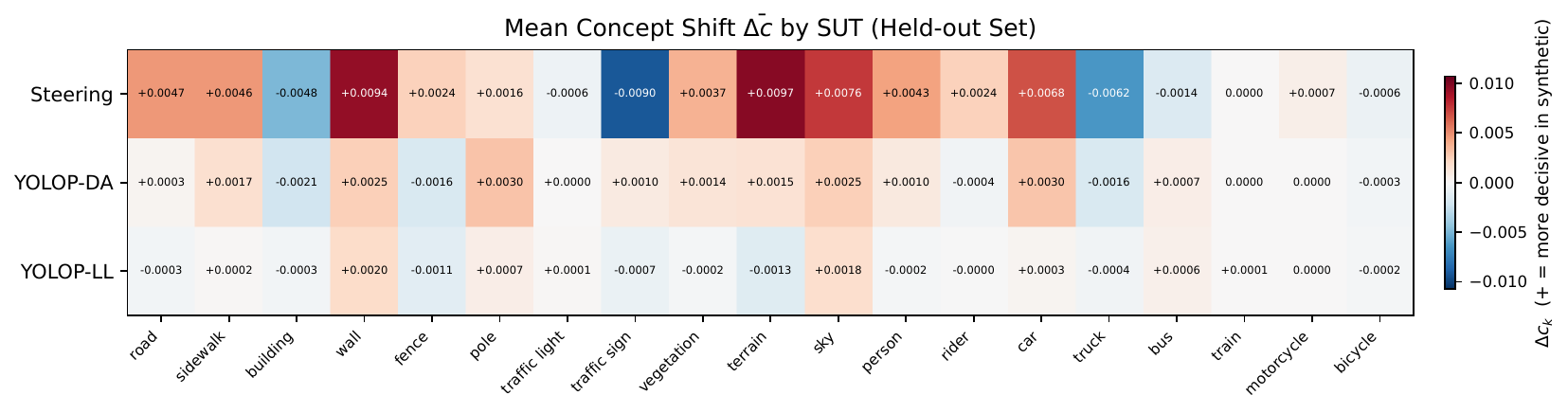}
\caption{Mean concept shift $\bar{\Delta\cvec}$ by SUT and concept. Positive values indicate that a concept becomes more decisive in synthetic imagery, negative values indicate that it becomes less decisive. Table~\ref{tab:top_shifts} lists the largest shifts numerically for readability.}
\label{fig:concept_shift}
\end{figure*}

\paragraph*{Concept-shift diagnostics}
Table~\ref{tab:top_shifts} and Figure~\ref{fig:concept_shift} summarize mean concept shifts $\bar{\Delta\cvec}$.
Across SUTs, building becomes less decisive for both YOLOP heads.
For Steering, the strongest positive mean shifts are terrain, wall, and sky, whereas traffic sign, truck, and building decrease.
For YOLOP--DA, car, pole, and wall increase most, while building, fence, and truck decrease.
For YOLOP--LL, wall and sky increase, whereas terrain, fence, and traffic sign decrease.
These ranked shifts yield concrete semantic targets for simulator and synthetic-data debugging.

\begin{table}[t]
\centering
\scriptsize
\caption{Top concept shifts per SUT. Each cell lists concept names with signed magnitudes from the dataset-level mean shift $\bar{\Delta\cvec}$.}
\label{tab:top_shifts}
\setlength{\tabcolsep}{3pt}
\renewcommand{\arraystretch}{1.00}
\begin{tabular}{@{}l p{0.38\columnwidth} p{0.38\columnwidth}@{}}
\toprule
\textbf{SUT} & \textbf{Top-$k$ increases (synthetic more decisive)} & \textbf{Top-$k$ decreases (synthetic less decisive)} \\
\midrule
Steering  & terrain (+0.0097), wall (+0.0094), sky (+0.0076) & traffic sign (-0.0090), truck (-0.0062), building (-0.0048) \\
YOLOP--DA & car (+0.0030), pole (+0.0030), wall (+0.0025) & building (-0.0021), fence (-0.0016), truck (-0.0016) \\
YOLOP--LL & wall (+0.0020), sky (+0.0018), pole (+0.0007) & terrain (-0.0013), fence (-0.0011), traffic sign (-0.0007) \\
\bottomrule
\end{tabular}
\end{table}

\paragraph*{Qualitative panels}
Figure~\ref{fig:qualitative} provides illustrative per-pair diagnostics.
Low-C-DFF exemplars typically preserve decisiveness over road-aligned concepts, while high-C-DFF exemplars show redistributed decisiveness toward non-transferable or over-emphasized concepts.
These panels are intended as semantic triage aids rather than primary evidence.

\ifshowqual
\begin{figure*}[!b]
\centering
\MaybeIncludeGraphics[
  width=0.98\textwidth,
  height=0.62\textheight,
  keepaspectratio
]{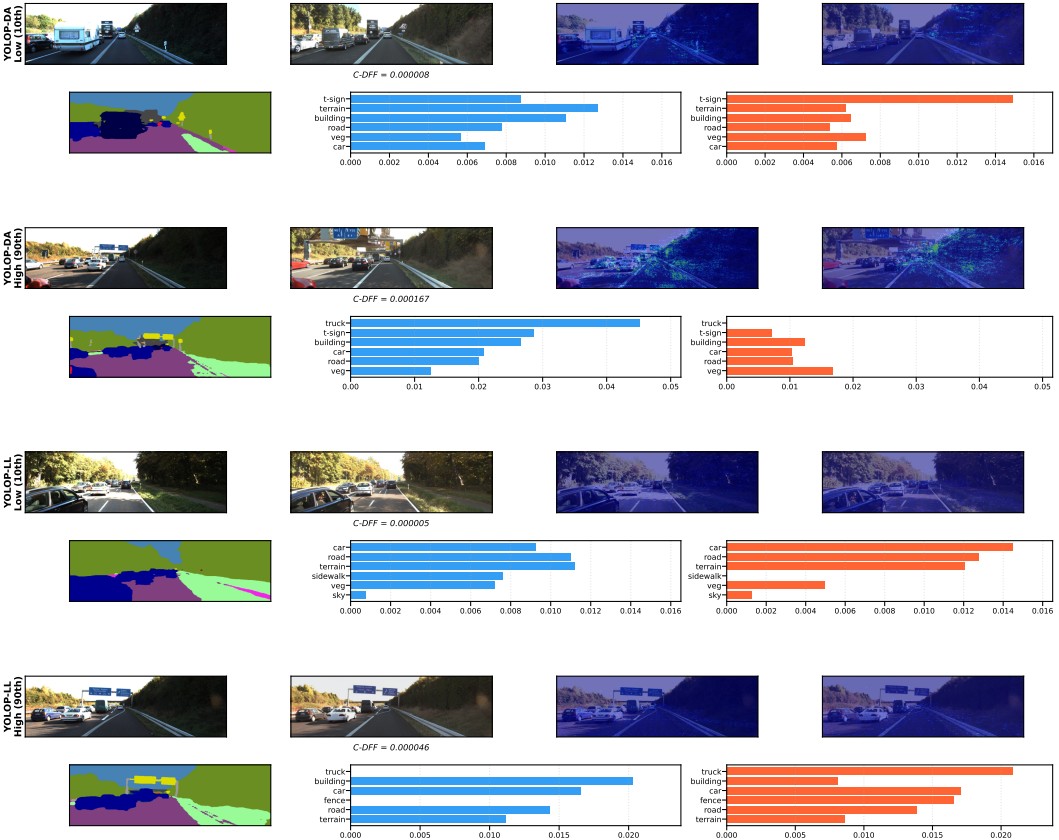}
\caption{Illustrative qualitative diagnostics. For each exemplar, we show real and synthetic inputs, decisive overlays, semantic segmentation, and top-6 concept bars. These selected examples are semantic triage aids rather than prevalence estimates. Quantitative results are reported in Tables~\ref{tab:main_results}--\ref{tab:top_shifts}.}
\label{fig:qualitative}
\end{figure*}
\fi

\section{Discussion}
\label{sec:discussion}

\subsection{Operational interpretation and workflow}
C-DFF and output-value (OV) similarity are not substitutes.
OV measures whether the SUT output remains similar across real and synthetic inputs, while C-DFF measures whether the SUT relies on similar semantic evidence.
Because OV is task-specific, we interpret it only within SUT: substantial output agreement can coexist with nontrivial mechanism shift, which is exactly the validation gap motivating mechanism-parity evaluation.

The concept-shift diagnostics make C-DFF operational rather than purely descriptive.
Held-out mean shifts identify concepts that are consistently over- or under-emphasized in synthetic imagery for a given SUT, giving simulator, renderer, or synthetic-data debugging a ranked semantic target list rather than a raw heatmap mismatch.

\paragraph*{When to use DFF vs.\ C-DFF}
DFF and C-DFF serve different purposes.
DFF is more suitable when the objective is decisive-map comparison itself or when decisive-feature distances are fed into a calibration loop \cite{Safaei2025DFF}.
C-DFF is more suitable when the objective is to explain \emph{why} a matched synthetic sample fails mechanism parity in a form actionable for debugging and V\&V triage.
They can be used together: DFF as a fine-grained mechanism signal, C-DFF as the semantic interpretation layer.

\label{sec:discussion_req}
In practice, C-DFF can be used in two layers:
(i) a scalar thresholded score for screening or regression testing, and
(ii) concept-level checks for debugging and requirement-style monitoring of critical concepts such as road, lane, and sidewalk.
The threshold calibration is scene-dependent, so we treat it as an operational summary for a given benchmark split rather than a universal bound.

\begin{table}[t]
\centering
\scriptsize
\caption{Practical C-DFF protocol for paired real--synthetic validation. The output is a scalar gate plus a ranked diagnostic of concept shifts.}
\label{tab:protocol}
\setlength{\tabcolsep}{3pt}
\renewcommand{\arraystretch}{1.00}
\begin{tabularx}{\columnwidth}{@{}lX@{}}
\toprule
\textbf{Step} & \textbf{What you do / what you log} \\
\midrule
1. Pairing & Construct $(x_r,x_s)$ with tight correspondence. Log pairing quality flags. \\
2. Scalar gate & Compute C-DFF and compare to calibrated $\eps$ for regression screening. Log pass rate. \\
3. Diagnose & Compute $\Delta \cvec$ and report top-$k$ increases and decreases per SUT for debugging. \\
4. Requirement-style checks & Track envelopes for critical concepts and alarm on drift beyond tolerances. \\
5. Act & Map dominant shifted concepts to simulator or data-generation knobs and re-test. \\
\bottomrule
\end{tabularx}
\end{table}

\subsection{Threats to validity and practical constraints}
\label{sec:threats}

\paragraph*{Decisive-map estimator variability}
C-DFF inherits variability from perturbation-based CF-XAI.
Although we use fixed optimization settings, clamping, and pooling, decisive maps can still vary with hyperparameters and random seeds \cite{Adebayo2018Sanity}.
We therefore treat the decisive-map extractor as a modular approximation, not as a ground-truth causal oracle.

\paragraph*{Segmentation as an interpretability scaffold}
\label{sec:threats_seg}
Concept-importance vectors depend on the semantic segmenter.
Systematic segmentation errors or real--synthetic domain shift can bias concept scores and inflate concept shifts for affected classes.
We partially bound this effect on the synthetic side by comparing Mask2Former predictions with VKITTI2 ground-truth labels for a held-out subset, but this remains one-sided because equivalent real-side ground-truth labels are unavailable.

\paragraph*{Pairing quality and correspondence}
\label{sec:threats_pairing}
C-DFF is defined on matched real--synthetic pairs.
If correspondence is imperfect (pose, timing, or object-layout mismatch), the score conflates domain effects with pairing noise.
This is mitigated in KITTI $\leftrightarrow$ Virtual~KITTI~2 by cloned trajectories, but pair-quality filtering remains important for future datasets.

\paragraph*{Compute cost}
Perturbation-based CF-XAI is computationally expensive.
Caching decisive maps and segmentations is essential for scaling C-DFF to larger benchmarks or additional SUTs.

\section{Conclusion}
\label{sec:conclusion}

C-DFF provides a semantic diagnostic layer for paired real--synthetic validation of frozen autonomous-driving perception models.
Instead of asking only whether matched inputs produce similar outputs, it asks whether decisiveness is distributed over the same scene concepts.
On KITTI $\leftrightarrow$ Virtual~KITTI~2, C-DFF differentiated the evaluated SUTs and exposed concept-level shifts useful for validation triage and simulator debugging.
Future work should test robustness to alternative decisive-map estimators, segmenters, pair-quality filters, and datasets.


\begin{thebibliography}{99}
\footnotesize
\bibitem{Riedmaier2020Survey}
S.~Riedmaier \emph{et al.}, ``Survey on scenario-based safety assessment of automated vehicles,'' \emph{IEEE Access}, 2020.

\bibitem{Zhong2021ScenarioSurvey}
Z.~Zhong, Y.~Tang, Y.~Zhou, V.~de~Oliveira~Neves, Y.~Liu, and B.~Ray, ``A survey on scenario-based testing for automated driving systems in high-fidelity simulation,'' arXiv:2112.00964, 2021.

\bibitem{Cheng2024SAFidelity}
C.-H.~Cheng, P.~St\"{o}ckel, and X.~Zhao, ``Instance-level safety-aware fidelity of synthetic data and its calibration,'' \emph{arXiv preprint} arXiv:2402.07031, 2024.

\bibitem{Safaei2025DFF}
D.~Safaei \emph{et al.}, ``Quantifying fidelity: A decisive feature approach to comparing synthetic and real imagery,'' arXiv:2512.16468, 2025.

\bibitem{Lapuschkin2019CleverHans}
S.~Lapuschkin, S.~W{\"a}ldchen, A.~Binder, G.~Montavon, W.~Samek, and K.-R.~M{\"u}ller, ``Unmasking Clever Hans predictors and assessing what machines really learn,'' \emph{Nature Communications}, vol.~10, art.~1096, 2019, doi:10.1038/s41467-019-08987-4.

\bibitem{Geirhos2020Shortcut}
R.~Geirhos, J.-H.~Jacobsen, C.~Michaelis, R.~Zemel, W.~Brendel, M.~Bethge, and F.~A.~Wichmann, ``Shortcut learning in deep neural networks,'' \emph{Nature Machine Intelligence}, vol.~2, pp.~665--673, 2020, doi:10.1038/s42256-020-00257-z.

\bibitem{FongVedaldi2017MeaningfulPerturbation}
R.~Fong and A.~Vedaldi, ``Interpretable explanations of black boxes by meaningful perturbation,'' in \emph{Proc.\ ICCV}, 2017.

\bibitem{Rudin1992TV}
L.~I.~Rudin, S.~Osher, and F.~Fatemi, ``Nonlinear total variation based noise removal algorithms,'' \emph{Physica D: Nonlinear Phenomena}, vol.~60, no.~1--4, pp.~259--268, 1992, doi:10.1016/0167-2789(92)90242-F.

\bibitem{Adebayo2018Sanity}
J.~Adebayo, J.~Gilmer, M.~Muelly, I.~Goodfellow, M.~Hardt, and B.~Kim, ``Sanity checks for saliency maps,'' in \emph{Proc.\ NeurIPS}, 2018.

\bibitem{Kim2018TCAV}
B.~Kim, M.~Wattenberg, J.~Gilmer, C.~Cai, J.~Wexler, F.~Viegas, and R.~Sayres, ``Interpretability beyond feature attribution: Quantitative testing with concept activation vectors (TCAV),'' in \emph{Proc.\ ICML}, 2018.

\bibitem{Koh2020CBM}
P.~W.~Koh, T.~Nguyen, Y.~S.~Tang, S.~Mussmann, E.~Pierson, B.~Kim, and P.~Liang, ``Concept bottleneck models,'' in \emph{Proc.\ ICML}, 2020.

\bibitem{Cheng2022Mask2Former}
B.~Cheng, I.~Misra, A.~G.~Schwing, A.~Kirillov, and R.~Girdhar, ``Masked-attention mask transformer for universal image segmentation,'' in \emph{Proc.\ CVPR}, 2022.

\bibitem{Cordts2016Cityscapes}
M.~Cordts \emph{et al.}, ``The Cityscapes dataset for semantic urban scene understanding,'' in \emph{Proc.\ CVPR}, 2016.

\bibitem{Wu2022YOLOP}
D.~Wu, M.-W.~Liao, W.-T.~Zhang, X.-G.~Wang, X.~Bai, W.-Q.~Cheng, and W.-Y.~Liu, ``YOLOP: You only look once for panoptic driving perception,'' \emph{Machine Intelligence Research}, vol.~19, no.~6, pp.~550--562, 2022, doi:10.1007/s11633-022-1339-y.

\bibitem{Bojarski2016E2E}
M.~Bojarski \emph{et al.}, ``End to End Learning for Self-Driving Cars,'' arXiv:1604.07316, 2016.

\bibitem{Chernikova2019Udacity}
A.~Chernikova, A.~Oprea, C.~Nita-Rotaru, and B.~Kim, ``Are Self-Driving Cars Secure? Evasion Attacks against Deep Neural Networks for Steering Angle Prediction,'' in \emph{Proc.\ IEEE SafeThings}, 2019.

\bibitem{Udacity2016Challenge2}
Udacity, ``Challenge 2: Using Deep Learning to Predict Steering Angles,'' 2016. [Online]. Available: \url{https://medium.com/udacity/challenge-2-using-deep-learning-to-predict-steering-angles-f42004a36ff3}. Accessed: Feb. 20, 2026.

\bibitem{Akiba2019Optuna}
T.~Akiba, S.~Sano, T.~Yanase, T.~Ohta, and M.~Koyama, ``Optuna: A Next-generation Hyperparameter Optimization Framework,'' in \emph{Proc.\ KDD}, 2019, doi:10.1145/3292500.3330701.

\bibitem{Borgefors1988Chamfer}
G.~Borgefors, ``Hierarchical chamfer matching: A parametric edge matching algorithm,'' \emph{IEEE Trans.\ Pattern Analysis and Machine Intelligence}, vol.~10, no.~6, pp.~849--865, 1988, doi:10.1109/34.9107.

\bibitem{Geiger2013KITTI}
A.~Geiger, P.~Lenz, C.~Stiller, and R.~Urtasun, ``Vision meets robotics: The KITTI dataset,'' \emph{The International Journal of Robotics Research}, vol.~32, no.~11, pp.~1231--1237, 2013, doi:10.1177/0278364913491297.

\bibitem{Cabon2020VKITTI2}
Y.~Cabon, N.~Murray, and M.~Humenberger, ``Virtual KITTI 2,'' arXiv:2001.10773, 2020.

\bibitem{Zhang2018LPIPS}
R.~Zhang, P.~Isola, A.~A.~Efros, E.~Shechtman, and O.~Wang, ``The unreasonable effectiveness of deep features as a perceptual metric,'' in \emph{Proc.\ CVPR}, 2018.

\bibitem{HFMask2FormerCityscapes}
Hugging Face, ``facebook/mask2former-swin-large-cityscapes-semantic (model card),'' [Online]. Available: \url{https://huggingface.co/facebook/mask2former-swin-large-cityscapes-semantic}. Accessed: Feb. 20, 2026.

\bibitem{KingmaBa2015Adam}
D.~P.~Kingma and J.~Ba, ``Adam: A Method for Stochastic Optimization,'' in \emph{Proc.\ ICLR}, 2015.

\bibitem{ADWGSTV2Whitepaper}
IEEE SA Autonomous Driving Working Group, ``Simulation, Testing, Verification, and Validation (STV2) of Autonomous Driving,'' White Paper, Oct.~2024. [Online]. Available: \url{https://sagroups.ieee.org/adwg/wp-content/uploads/sites/661/2024/10/ADWG_STV2_whitepaper.pdf}.

\end{thebibliography}
\end{document}